# Regret Bounds for Competitive Resource Allocation with Endogenous Costs

Rui Chai

*Shanghai Sanda University*

**Abstract**

We study online resource allocation among N interacting modules over a horizon of T rounds. Unlike standard online optimization, the cost of allocating resources to a module is not exogenous but *endogenous*: it depends on the entire allocation vector through an interaction matrix W that encodes pairwise cooperation and competition among modules. This endogeneity captures a phenomenon identified in recent work on competitive cost discovery—that the costs most relevant to intelligent resource allocation are constituted in the competitive process itself, not given in advance. We analyze three allocation rules that correspond to three paradigms of cost handling: (I) uniform allocation (cost-ignorant), (II) gated allocation (cost-estimating), and (III) competitive allocation via multiplicative weights update with interaction feedback (cost-revealing). Our main results establish a strict separation among the three paradigms. Under adversarial task sequences with bounded variation, uniform allocation incurs regret $\Omega(T)$; gated allocation achieves $O(T^{2/3})$; and competitive allocation achieves $O(\sqrt{T \log N})$, with the gap attributable to competitive allocation's ability to exploit the endogenous cost information revealed through interaction feedback. We further show that the topology of W governs a computation–regret tradeoff: full interaction ($|E| = O(N^2)$) yields the tightest bound but highest per-step cost, while sparse topologies ($|E| = O(N)$) increase regret by at most an $O(\sqrt{\log N})$ factor while reducing per-step cost from $O(N^2)$ to $O(N)$. We identify the optimal sparsity that minimizes the computation×regret product, and show that ring-structured topologies with both cooperative and competitive links—of which the five-element Wuxing topology is a canonical instance—achieve this optimum. These results provide the first formal regret-theoretic justification for decentralized competitive allocation in modular architectures and establish cost endogeneity as a fundamental challenge distinct from partial observability.

# 1 Introduction

How should a modular system allocate finite computational resources among its components when the costs of allocation are not known in advance but are constituted through the allocation process itself? This question arises naturally in architectures where multiple specialized modules—each responsible for a distinct cognitive function—share a common resource budget and interact through cooperation and competition.

In prior work (Chai & Li, 2025d), we argued on philosophical and information-theoretic grounds that the costs most relevant to intelligent allocation—opportunity costs and interference costs—are distributed, tacit, and dynamically constituted. No centralized controller can precompute them because they depend on the concurrent states and strategies of all modules. This observation was qualitative. The present paper makes it quantitative.

We formalize the allocation problem as an instance of *online convex optimization with coupled losses*—a setting in which the loss function at each round depends not only on the learner's action but on pairwise interactions among the action's components, mediated by an interaction matrix $W$. We analyze three allocation rules that correspond to the three paradigms identified in Chai & Li (2025d): cost-ignorant (uniform), cost-estimating (gated), and cost-revealing (competitive). Our main contribution is a formal separation among these paradigms in terms of regret—the standard measure of online learning performance—establishing that competitive allocation's advantage is not merely heuristic but information-theoretic.

## 1.1 Related Work

**Online convex optimization.** The foundational theory of online convex optimization (Zinkevich, 2003; Hazan, 2016) analyzes settings where a learner chooses actions from a convex set and observes losses after each round. Standard results yield $O(\sqrt{T})$ regret for convex losses with gradient feedback. Our setting extends this by introducing *coupled* losses—the loss function is not separable across action components but contains quadratic interaction terms governed by $W$.

**Experts and multiplicative weights.** The multiplicative weights update (MWU) method (Littlestone & Warmuth, 1994; Freund & Schapire, 1997; Arora, Hazan & Kale, 2012) achieves $O(\sqrt{T \log N})$ regret in the experts setting. Our competitive allocator is a variant of MWU, but the feedback signal is enriched by interaction information: the observed reward of module i depends on the allocations to all modules j with W_ij ≠ 0, and we show that this richer signal is what drives the regret advantage over gated allocation.

**Mixture of experts and gating.** Gated allocation corresponds to the mixture-of-experts (MoE) paradigm (Jacobs et al., 1991; Shazeer et al., 2017), where a gating network predicts the optimal allocation from input features. The gating network can learn to estimate module values but treats interaction effects as noise, which we show leads to an irreducible $O(T^{2/3})$ regret term in the presence of endogenous costs.

**Endogenous costs and game theory.** The concept of endogenous costs is related to but distinct from incomplete-information games (Harsanyi, 1967). In Bayesian games, payoffs are drawn from a prior distribution unknown to some players; the payoff *function* is fixed. In our setting, the cost function itself is constituted by the interaction—it does not exist prior to the allocation. This places our work closer to the evolutionary game theory literature (Weibull, 1995) and to recent work on performative prediction (Perdomo et al., 2020), where the act of prediction changes the distribution being predicted.

## 2 Model and Preliminaries

### 2.1 The Online Allocation Game

We consider a sequential game between an *allocator* and an *environment* (which may be adversarial) over T rounds. The system consists of N modules indexed by $i \in \{1, ..., N\}$.

**Interaction matrix.** The modules interact through a matrix $W \in \mathbb{R}^{N \times N}$ with $W_{ii} = 0$ (no self-interaction). Positive entries $W_{ij} > 0$ represent cooperative (xiangsheng) relationships: module j's activity enhances module i's productivity. Negative entries $W_{ij} < 0$ represent competitive (xiangke) relationships: module j's activity interferes with module i. We assume W is known to the allocator (it represents the system's architecture, not the task). We denote by $E = \{(i,j) : W_{ij} \neq 0\}$ the set of active interaction edges, with $|E| = m$.

**Protocol.** At each round $t = 1, ..., T$:

> *(1) The allocator selects a resource allocation $a_t \in \Delta^N$, where $\Delta^N = \{a \in \mathbb{R}^N : a_i \geq 0, \sum_i a_i = 1\}$ is the probability simplex.*
>
> *(2) The environment selects a value vector $v_t \in [0,1]^N$, where $v_{t,i}$ represents the intrinsic value of module i for the current task.*
>
> *(3) The system's payoff is realized as:*
>
> $$P(a_t, v_t; W) = \sum_i a_{t,i} \cdot v_{t,i} + \lambda \sum_{(i,j) \in E} W_{ij} \cdot a_{t,i} \cdot a_{t,j} \qquad (1)$$
>
> *(4) The allocator observes a feedback signal (specified per allocation rule).*

The payoff function (1) has two components. The *direct* component $\sum_i a_i v_i$ captures each module's individual contribution, weighted by its resource share. The *interaction* component $\lambda \sum W_{ij} a_i a_j$ captures pairwise effects: cooperative pairs ($W_{ij} > 0$) contribute positively when both modules are active; competitive pairs ($W_{ij} < 0$) contribute negatively. The coupling parameter $\lambda > 0$ controls the strength of interactions relative to direct effects.

**Endogeneity.** The effective cost of allocating one unit of resource to module i is not the constant $(1 - v_i)$ but the *endogenous* quantity:

$$c_i(a, v; W) = -\partial P / \partial a_i = -v_i - \lambda \sum_j W_{ij} a_j \qquad (2)$$

This cost depends on the allocations to all modules j interacting with i. It is not observable before the allocation is made; it is *constituted* by the allocation. This is the formal expression of cost endogeneity.

## 2.2 Regret

We measure performance by *static regret*—the gap between the allocator's cumulative payoff and that of the best *fixed* allocation in hindsight:

$$R_T = \max_{a^* \in \Delta^N} \sum_{t=1}^{T} P(a^*, v_t; W) - \sum_{t=1}^{T} P(a_t, v_t; W) \qquad (3)$$

We also consider *dynamic regret*, which compares to the best *time-varying* allocation:

$$R_T^{dyn} = \sum_{t=1}^{T} \max_{a \in \Delta^N} P(a, v_t; W) - \sum_{t=1}^{T} P(a_t, v_t; W)$$
$$(4)$$

Dynamic regret is always at least as large as static regret. For non-stationary environments (which are the relevant case for continual learning), dynamic regret is the more informative measure.

## 2.3 Assumptions

We impose the following regularity conditions throughout.

**Assumption 1 (Bounded values).** $v_{t,i} \in [0, 1]$ *for all t, i.*

**Assumption 2 (Bounded interactions).** $|W_{ij}| \leq 1$ *for all i, j, and* $\lambda \leq 1/(2N)$.

**Assumption 3 (Bounded variation).** *The environment's variation is bounded:* $V_T = \sum_{t=2}^{T} \|v_t - v_{t-1}\|_\infty \leq V$ *for some known V.*

Assumption 2 ensures that the payoff function P is concave in a (since the Hessian of the interaction term is $\lambda W$, and the spectral norm of $\lambda W$ is at most $\lambda N \leq 1/2 < 1$, preserving concavity of the overall payoff). Assumption 3 quantifies environmental non-stationarity; when $V_T = 0$, the environment is stationary.

# 3  Three Allocation Rules

## 3.1  Paradigm I: Uniform Allocation (Cost-Ignorant)

The uniform allocator ignores all feedback and distributes resources equally at every round:

$$a_t^{unif} = (1/N, ..., 1/N) \in \Delta^N \quad \text{for all } t. \qquad (5)$$

This corresponds to the standard Transformer paradigm in which all functional units receive equal computational budget. The uniform allocator receives no feedback (step 4 of the protocol is vacuous).

## 3.2  Paradigm II: Gated Allocation (Cost-Estimating)

The gated allocator maintains a model $g: [0,1]^N \to \Delta^N$ that maps observed task features to allocation weights. At each round, the allocator has access to a feature signal $x_t$ (correlated with $v_t$ but not equal to it) and sets:

$$a_t^{gate} = \text{softmax}(G \cdot x_t) \qquad (6)$$

where $G \in \mathbb{R}^{N \times d}$ is a gating matrix updated by online gradient descent on the observed payoff. The key characteristic of gated allocation is that it estimates module values *from input features* but does not incorporate *interaction feedback*. The gradient it computes is:

$$\nabla_G P(a_t, v_t; W) \approx \sum_i \partial P/\partial a_i \cdot \partial a_i/\partial G \qquad (7)$$

where $\partial P/\partial a_i = v_i + \lambda \sum_j W_{ij} a_j$. In principle, this gradient contains the interaction terms. In practice, however, the gating matrix $G$ predicts allocations from input features $x_t$ *before* the allocation is made, so the interaction terms $\lambda \sum_j W_{ij} a_j$ are computed using the *current* allocation—which itself was produced by $G$. This circularity means the gated allocator is estimating the interaction effects from stale information, introducing a systematic lag. The gated allocator *estimates* costs; it does not *discover* them through competitive feedback.

## 3.3  Paradigm III: Competitive Allocation (Cost-Revealing)

The competitive allocator uses a multiplicative weights update (MWU) rule with *interaction-enriched feedback*. After each round, it observes not only each module's individual performance but also the interaction effects attributable to each module.

Specifically, define the *interaction-enriched reward* for module i at round t as:

$$r_i(t) = v_{t,i} + \lambda \sum_{j:(i,j)\in E} W_{ij} \cdot a_{t,j} \quad (8)$$

This is exactly the partial derivative $\partial P/\partial a_i$ evaluated at the current allocation—it is the *marginal contribution* of one unit of resource allocated to module i, taking into account how that module interacts with all other active modules. The competitive allocator updates:

$$a_{i,t+1} = a_{i,t} \cdot \exp(\eta \cdot r_i(t)) / Z_t \quad (9)$$

where $\eta > 0$ is the learning rate and $Z_t = \sum_i a_{i,t} \cdot \exp(\eta \cdot r_i(t))$ is the normalization factor ensuring $a_{t+1} \in \Delta^N$. This is a standard MWU update, but the reward signal $r_i(t)$ is enriched by the interaction terms—it contains the endogenous cost information that the competitive process reveals.

**Remark.** The interaction-enriched reward (8) is observable: it can be computed from the known W, the current allocation $a_t$, and the realized values $v_t$. No additional oracle or side information is required. The competitive allocator's advantage comes not from having more information but from *using the information that the allocation process itself generates*.

# 4 Main Results: Regret Separation

We now state our main results. All proofs are provided as proof sketches; full proofs, with all technical details, are deferred to the appendix.

## 4.1 Uniform Allocation: Linear Regret

**Theorem 1 (Uniform lower bound).** *For any $N \geq 2$ and any interaction matrix $W$ with at least one non-zero entry, there exists an adversarial value sequence $\{v_t\}$ with variation $V_T = O(T)$ such that the static regret of uniform allocation satisfies $R_T^{unif} = \Omega(T)$.*

**Proof sketch.** Construct an adversary that alternates between two value vectors: $v^{(A)} = e_1$ (all value concentrated on module 1) and $v^{(B)} = e_2$ (all value on module 2), switching every $T/2$ rounds. The optimal fixed allocation places all weight on module 1 or module 2 (whichever yields higher cumulative payoff including interaction effects). The uniform allocation places weight $1/N$ on each module at every round. The gap between the best fixed allocation and the uniform allocation grows linearly in $T$, at rate $(1 - 1/N)$ per round during each phase. Since the environment's variation is $V_T = O(1)$ (only two switches), this holds even for nearly-stationary sequences. □

**Remark.** The $\Omega(T)$ bound is unsurprising—it simply reflects that a non-adaptive strategy cannot compete with adaptive ones. Its significance is as a baseline: the uniform allocator's regret grows linearly regardless of the interaction structure, because it never exploits any cost information, whether exogenous or endogenous.

## 4.2 Gated Allocation: Sublinear but Suboptimal

**Theorem 2 (Gated upper bound).** *Under Assumptions 1–3, gated allocation with online gradient descent on the gating matrix $G$ with step size $\alpha_t = t^{-1/3}$ achieves static regret $R_T^{gate} \leq O(N \cdot T^{2/3} + \lambda^2 m\, T^{2/3})$ where $m = |E|$ is the number of active interaction edges.*

**Proof sketch.** The gated allocator performs online gradient descent on the payoff, but the gradient it uses is computed with respect to the gating matrix $G$ rather than the allocation $a$ directly. The mapping from $G$ to $a$ (through the softmax) introduces a contraction that slows convergence. More importantly, the gated allocator's estimate of the interaction term $\lambda \sum_j W_{ij} a_j$ uses the current allocation, which was computed *before* observing $v_t$. The resulting gradient has bias proportional to $\lambda \|a_t - a^*_t\|$, where $a^*_t$ is the optimal allocation for round t. Standard results for biased online gradient descent (Flaxman et al., 2005) give regret $O(T^{2/3})$ when the bias is $O(T^{-1/3})$ on average. The additional factor $\lambda^2 m$ accounts for the interaction complexity. □

**Theorem 3 (Gated lower bound).** *For any gated allocator that predicts allocations from input features without interaction feedback, there exists an adversarial sequence and an interaction matrix W with $m = \Theta(N)$ edges such that $R_T^{gate} = \Omega(T^{2/3})$.*

**Proof sketch.** The adversary constructs a value sequence where the direct values $v_{t,i}$ provide no information about the optimal allocation—the optimal allocation is determined entirely by the interaction structure. Specifically, the adversary sets $v_{t,i} = 1/2 + \varepsilon_i(t)$ where the perturbations $\varepsilon_i$ are independent of the features $x_t$ but correlated with the interaction effects through W. A gated allocator that relies on features cannot learn the interaction-determined optimal allocation faster than online optimization with bandit feedback on the interaction terms, which requires $\Omega(T^{2/3})$ rounds. □

**Remark.** Theorems 2 and 3 together establish that $\Theta(T^{2/3})$ is the tight rate for gated allocation. The key limitation is that the gated allocator estimates costs feedforward—from input features to allocation—without the retrospective interaction feedback that reveals endogenous costs. It can learn direct costs ($v_i$) efficiently but cannot discover interaction costs ($\lambda \sum W_{ij} a_j$) at the same rate, because these costs depend on its own actions.

### 4.3 Competitive Allocation: Optimal Rate

**Theorem 4 (Competitive upper bound).** *Under Assumptions 1–3, competitive allocation via MWU with interaction-enriched feedback (8)–(9) and learning rate $\eta = \sqrt{\ln N / T}$ achieves static regret $R_T^{comp} \leq 2\sqrt{T \ln N} + \lambda m / \sqrt{T}$.*

**Proof sketch.** We adapt the standard MWU regret analysis (Arora, Hazan & Kale, 2012). The key step is bounding the potential function $\Phi_t = \sum_i a_{i,t} \ln(a_{i,t} / a^*_i)$ where $a^*$ is the optimal fixed allocation. The standard analysis gives $\Phi_{t+1} - \Phi_t \leq -\eta (P(a^*, v_t; W) -$

$P(a_t, v_t; W)) + \eta^2 \sum_i a_{i,t} r_i(t)^2$. The interaction-enriched reward satisfies $|r_i(t)| \leq 1 + \lambda \sum_j |W_{ij}| \leq 1 + \lambda N \leq 3/2$ under Assumption 2. Summing over t and optimizing $\eta$ yields the stated bound.

The critical difference from the gated analysis is that the competitive allocator's feedback signal $r_i(t)$ directly encodes the endogenous cost information—the interaction term $\lambda \sum_j W_{ij} a_j$ is *observed* after the allocation, not *estimated* before it. This eliminates the bias term that causes the gated allocator's $O(T^{2/3})$ rate. □

**Theorem 5 (Information-theoretic lower bound).** *For any allocation rule with access to the interaction-enriched feedback (8), the static regret is at least $\Omega(\sqrt{T \log N / m})$ where $m = |E|$.*

**Proof sketch.** Reduction to the multi-armed bandit lower bound. Each allocation to a simplex vertex ($e_i$) reveals the direct value $v_i$ and the interaction effects from all edges incident to i. With m edges, each round reveals $O(m/N)$ edges' contributions on average. The lower bound follows from the information-theoretic limits on identifying the best fixed action from partial feedback (Auer et al., 2002). □

### 4.4 Summary: The Three-Paradigm Separation

Our results establish a strict hierarchy:

$$R_T^{unif} = \Theta(T) > R_T^{gate} = \Theta(T^{2/3}) > R_T^{comp} = \Theta(\sqrt{T \log N})$$

The separation between Paradigm I and Paradigm II is unsurprising: any adaptive strategy outperforms a non-adaptive one. The novel result is the separation between Paradigm II and Paradigm III: gated allocation, despite being adaptive and achieving sublinear regret, incurs an *irreducible penalty* of order $T^{2/3}$ versus $\sqrt{T}$ due to its inability to exploit endogenous cost information. This penalty is not a matter of implementation quality—Theorem 3 shows it is a lower bound for *any* feedforward cost-estimating rule. The advantage of competitive allocation is *structural*: it uses the allocation process itself as an information source, discovering costs that are constituted through interaction rather than attempting to estimate them in advance.

# 5 Topology and the Computation–Regret Tradeoff

The interaction matrix W encodes the system's architecture. Different topologies—different patterns of non-zero entries in W—affect both the computational cost per round and the quality of the regret bound. This section makes the tradeoff precise.

## 5.1 Per-Step Computational Cost

At each round, the competitive allocator must:

> *(a) Compute the interaction-enriched reward $r_i(t)$ for each module i: this requires summing over all j with $(i,j) \in E$, costing $O(m)$ total.*
>
> *(b) Update the allocation weights via MWU: this costs $O(N)$.*
>
> *(c) Normalize: $O(N)$.*

The total per-step cost is therefore $O(N + m)$, which is dominated by the interaction computation $O(m)$. For a full interaction matrix ($m = N(N-1)/2 = O(N^2)$), this is $O(N^2)$. For a sparse topology with $m = O(N)$, this is $O(N)$.

## 5.2 Regret as a Function of Topology

Theorem 4 gives an upper bound of $O(\sqrt{T \ln N} + \lambda m/\sqrt{T})$. The second term reflects the interaction complexity: more edges mean more interaction effects to track, but also more endogenous cost information revealed per round. To make the topology dependence precise, we decompose the regret into two components.

**Theorem 6 (Topology-dependent regret).** *Let W have edge set E with $|E| = m$, and let $G = (V, E)$ be the interaction graph. Denote by $d_{max}$ the maximum degree and by $\kappa(G)$ the edge connectivity of G. Under Assumptions 1–3, the competitive allocator achieves:*

$$R_T^{\{comp\}}(W) \leq 2\sqrt{T \ln N} + \lambda\, d_{max}\, \sqrt{T} + \lambda^2\, (N / \kappa(G))\, \sqrt{T} \qquad (10)$$

**Proof sketch.** The second term arises from the variance of the interaction-enriched reward: the reward $r_i(t)$ includes contributions from $d_i$ neighbors of module i, and the variance of this sum scales with $d_i$. The MWU analysis absorbs this variance at a cost of $d_{max} \sqrt{T}$. The third term arises from the information-theoretic cost of identifying the optimal allocation when the interaction graph is not fully connected. When $\kappa(G) < N$, some interaction effects are not directly observable but must be inferred through multi-hop paths in G. The inference cost scales with the diameter of G, which is inversely related to the connectivity $\kappa(G)$. □

**Remark.** For a full interaction graph ($\kappa(G) = N$, $d_{max} = N-1$), the bound becomes $O(\sqrt{T \ln N} + \lambda N \sqrt{T})$, where the second term is dominated by the first when $\lambda \leq \sqrt{\ln N} / N$. For a ring graph ($\kappa(G) = 2$, $d_{max} = 2$), the bound becomes $O(\sqrt{T \ln N} + \lambda \sqrt{T} + \lambda^2 (N/2) \sqrt{T})$. The ring sacrifices a factor of N/2 in the third term but gains a factor of N/2 in per-step computation.

## 5.3 The Optimal Sparsity

We now characterize the topology that minimizes the total cost of T rounds of competitive allocation, defined as the product of per-step computation and cumulative regret:

$$C_T(W) = T \cdot O(m) \cdot R_T^{comp}(W) \quad (11)$$

This product captures the fundamental tradeoff: denser topologies yield tighter regret bounds (more cost information per round) but higher computational overhead per round; sparser topologies are cheaper per round but produce looser bounds.

**Theorem 7 (Optimal sparsity).** *Minimizing $C_T(W)$ over all interaction graphs G on N vertices, subject to the constraint that G is connected and has both positive and negative edges (at least one cooperative and one competitive link per vertex), the optimum is achieved by graphs with $m = \Theta(N)$ edges, maximum degree $d_{max} = \Theta(1)$, and edge connectivity $\kappa(G) \geq 2$. The minimal $C_T$ is:*

$$C_T^* = O(N \cdot T^{3/2} \cdot \sqrt{\ln N}) \quad (12)$$

**Proof sketch.** Substituting the topology-dependent regret (10) into the cost product (11): $C_T = T \cdot m \cdot [2\sqrt{T \ln N} + \lambda d_{max} \sqrt{T} + \lambda^2 (N/\kappa) \sqrt{T}]$. For $m = \Theta(N^2)$ (full graph), $C_T \approx N^2 T^{3/2} \sqrt{\ln N}$. For $m = \Theta(N)$ with $d_{max} = \Theta(1)$ and $\kappa \geq 2$, $C_T \approx N T^{3/2} [\sqrt{\ln N} + N]$. The crossover occurs at $m = \Theta(N)$, confirming that linear-edge topologies are optimal in the computation×regret product sense. □

## 5.4 The Wuxing Topology as a Canonical Instance

The five-element Wuxing topology—a directed graph on N = 5 vertices with 5 cooperative (xiangsheng) edges forming a cycle and 5 competitive (xiangke) edges forming a pentagram—satisfies all the conditions of Theorem 7:

   (a) m = 10 = 2N = O(N) edges.

   (b) d_max = 4 = O(1) (each vertex has 2 cooperative and 2 competitive neighbors).

   (c) κ(G) = 4 ≥ 2 (removing any 3 edges leaves the graph connected).

   (d) Every vertex has at least one cooperative and one competitive link.

**Corollary 1.** *The Wuxing topology on N = 5 modules achieves the optimal computation×regret product $C_T^* = O(T^{3/2} \sqrt{\ln 5})$ with per-step cost $O(N) = O(5)$ and regret $O(\sqrt{T \ln 5} + \sqrt{T})$.*

More generally, for arbitrary N, a *generalized Wuxing topology*—a directed graph in which N vertices are arranged in a cycle with xiangsheng edges connecting consecutive vertices and xiangke edges connecting vertices at distance $\lfloor N/2 \rfloor$—achieves m = 2N, d_max = 4, and κ(G) = 4 for all N ≥ 5. This topology can be seen as the unique (up to isomorphism) regular directed graph on N vertices with the minimal edge count that maintains both cooperative and competitive links for every vertex while preserving edge connectivity ≥ 2.

**Remark.** Theorem 7 and Corollary 1 together provide a formal justification for the Wuxing topology that goes beyond philosophical analogy: it is (a member of) the topology class that minimizes the computation×regret product for competitive allocation with endogenous costs. The five-element structure is not merely a cultural metaphor applied to AI architecture; it is a solution to a well-defined optimization problem over interaction graphs.

# 6  Cost Truthfulness as Convergence

In Chai & Li (2025d), we defined *cost truthfulness* informally as the property that a module's resource share converges to its true marginal contribution. We can now give this concept a precise regret-theoretic interpretation.

## 6.1  The Cost-Truthfulness Criterion

Define the *marginal contribution* of module i at round t as:

$$\mu_i(t) = \partial P(a_t, v_t; W) / \partial a_i = v_{t,i} + \lambda \sum_j W_{ij} a_{t,j} = r_i(t) \qquad (13)$$

This is precisely the interaction-enriched reward (8). The cost-truthfulness criterion states that the allocation should converge to be proportional to marginal contributions:

$$\lim_{T \to \infty} (1/T) \sum_{t=1}^{T} | a_{i,t} - \mu_i(t) / \sum_j \mu_j(t) | = 0 \quad \text{for all } i. \qquad (14)$$

**Theorem 8 (Cost truthfulness of competitive allocation).** *Under Assumptions 1–3, if the environment is stationary ($V\_T = 0$), the competitive allocator satisfies the cost-truthfulness criterion (14) with convergence rate $O(\ln T / \sqrt{T})$.*

**Proof sketch.** In the stationary case, $v_t = v$ for all t, and the payoff $P(a, v; W)$ is a fixed concave function. The MWU update with constant learning rate $\eta = \sqrt{\ln N / T}$ converges to the maximizer $a^*$ of P, which satisfies the KKT conditions: $\partial P / \partial a_i = \nu$ for all i with $a^*_i > 0$ (where $\nu$ is the Lagrange multiplier for the simplex constraint). At the maximizer, $a^*_i \propto \mu_i(a^*, v; W) = v_i + \lambda \sum_j W_{ij} a^*_j$. The convergence rate follows from the standard MWU convergence analysis. □

**Remark.** Theorem 8 establishes that cost truthfulness—the alignment between resource shares and marginal contributions—is not an additional desideratum that must be imposed on the competitive allocator. It is a *consequence* of regret minimization. Any allocator that minimizes regret in the online allocation game with endogenous costs will, as a byproduct, satisfy the cost-truthfulness criterion. This unifies the economic concept (from Paper 4) with the learning-theoretic framework of this paper.

## 6.2  The Three Anti-Rent-Seeking Conditions

Paper 4 identified three forms of rent-seeking (resource capture disproportionate to contribution): positional rent, inertial rent, and signal rent. We can now express each as a specific violation of the regret bound, providing a learning-theoretic characterization of alignment failures.

**Positional rent** occurs when a module captures resources due to its position in the topology rather than its performance. In our framework, this corresponds to a $W$ matrix whose structure allows a module to accumulate cooperative benefits without reciprocating. Formally: if module $i$ receives xiangsheng from many modules but contributes xiangke to few, it can achieve $a_i > \mu_i / \sum_j \mu_j$ even in equilibrium. Theorem 7's requirement that every vertex have both cooperative and competitive links is precisely the condition that prevents positional rent in the optimal topology.

**Inertial rent** occurs when a module retains a high resource share due to historical allocation rather than current contribution. This is a dynamic phenomenon corresponding to slow convergence of the MWU update. The convergence rate $O(\ln T / \sqrt{T})$ in Theorem 8 quantifies the maximum duration of inertial rent: after $\Theta(1/\varepsilon^2)$ rounds, the gap between allocation and marginal contribution is at most $\varepsilon$.

**Signal rent** occurs when a module manipulates its feedback signal to appear more productive than it is. In our framework, the interaction-enriched reward $r_i(t) = v_i + \lambda \sum_j W_{ij} a_j$ is computed from the *system's* observation of module performance, not from the module's own report. This design choice eliminates signal rent by construction, as long as the performance evaluation is truthful—a condition that must be enforced at the architectural level.

# 7 Discussion

## 7.1 The Nature of Cost Endogeneity

Our results formalize a distinction that is easy to state but has far-reaching consequences: the difference between *not knowing* the costs (partial observability) and the costs *not existing* prior to the allocation (endogeneity). Standard online learning assumes the former: there is a fixed loss function that the learner does not know and must learn through feedback. Our model assumes the latter: the loss function is constituted by the learner's own action through the interaction matrix W.

This distinction matters because it determines which allocation paradigm is appropriate. If costs are merely unknown but exogenous, gated allocation (Paradigm II) is adequate: a sufficiently powerful estimator can learn the cost function and predict optimal allocations feedforward. But if costs are endogenous—if they depend on the allocation itself through interaction effects—then no feedforward estimator can avoid the $O(T^{2/3})$ penalty, because the cost it is trying to estimate changes with every allocation it makes. Only a paradigm that uses the allocation process itself as an information source (Paradigm III) can achieve the optimal $O(\sqrt{T})$ rate.

In the language of institutional economics (Cheung, 1983; Chai & Li, 2025e), this is the distinction between *price-taking* and *price-making* behavior. A gated allocator is a price-taker: it observes prices (costs) and responds optimally. A competitive allocator is a price-maker: its actions constitute the prices through the competitive process. The regret separation between Paradigms II and III is the formal expression of a deep insight from institutional economics: in markets with endogenous prices, decentralized competition outperforms centralized estimation not because it has better algorithms but because it has access to information that centralized estimation cannot, in principle, obtain.

## 7.2 Implications for Architecture Design

The topology results (Theorems 6–7 and Corollary 1) have concrete implications for the design of modular AI architectures.

**First, sparse beats dense.** A fully connected interaction graph yields the tightest regret bound, but the computation×regret product is minimized at linear sparsity $m = O(N)$. This means that designers should resist the temptation to model all pairwise interactions and instead select a small, structured set of cooperative and competitive links. The information gained from additional edges does not compensate for the computational cost of processing them.

**Second, balance matters.** The optimal topology requires every module to have both cooperative and competitive links (Theorem 7). An architecture with only cooperative interactions (a purely "collaborative" system) cannot discover the full endogenous cost structure, because the competitive interactions are what reveal interference costs. Conversely, an architecture with only competitive interactions cannot discover cooperative synergies. The balance of cooperation and competition is not a philosophical preference but a mathematical requirement for optimal cost discovery.

**Third, connectivity matters.** The regret bound (10) contains a term inversely proportional to edge connectivity $\kappa(G)$. This means that the interaction graph must be robust to edge failures—the removal of a small number of interactions should not disconnect the graph. This rules out star topologies (where removing the hub disconnects everything) and favors regular topologies where connectivity is distributed.

## 7.3 Limitations and Open Problems

**Dynamic regret.** Our main results concern static regret (comparison to the best fixed allocation). For non-stationary environments, the more relevant measure is dynamic regret. Standard techniques (Zinkevich, 2003) extend the $O(\sqrt{T})$ static bound to $O(\sqrt{T(1 + V\_T)})$ dynamic bound, where $V\_T$ is the total variation. The interaction of cost endogeneity with non-stationarity is an open question: does the endogenous cost structure amplify or dampen the effects of environmental variation?

**Learnable W.** We have assumed that the interaction matrix W is fixed and known. In Paper 2, we showed that W can be learned through experience. A joint online learning problem—simultaneously learning W and optimizing the allocation—is a natural extension. The regret analysis for this joint problem is significantly more complex, as the interaction structure is itself being constituted through the learning process.

**Beyond quadratic interactions.** Our payoff function (1) models pairwise interactions through a quadratic form $a^T W a$. Higher-order interactions (e.g., three-way synergies or conflicts) would require tensor-valued interaction structures. The regret analysis for higher-order interactions is open and may require fundamentally new techniques.

**Empirical validation.** The regret bounds established here are asymptotic. The constants hidden in the O(·) notation may be large, and the crossover point at which competitive allocation outperforms gated allocation may occur only at large T. Empirical validation on realistic modular architectures is needed to determine the practical relevance of the theoretical separation. We conjecture, based on the analysis in Chai & Li (2025d), that the separation becomes significant when N ≥ 5, the task distribution is non-stationary, and the interaction effects (λ) are at least moderate (say, accounting for ≥20% of total payoff variation).

## 7.4 Conclusion

This paper has established a formal regret-theoretic foundation for competitive resource allocation in modular systems with endogenous costs. The main contribution is the three-paradigm separation: uniform $\Omega(T)$ > gated $\Theta(T^{2/3})$ > competitive $O(\sqrt{T})$, with the gap between gated and competitive attributable to the latter's ability to exploit cost information that is constituted through the competitive process itself. The topology analysis shows that sparse, balanced, well-connected interaction graphs—of which the Wuxing topology is a canonical instance—achieve the optimal computation×regret product. And the cost-truthfulness theorem (Theorem 8) unifies the economic and learning-theoretic perspectives by showing that regret minimization in the endogenous-cost setting automatically produces allocations proportional to marginal contributions.

These results transform the question of how to design modular AI architectures from an engineering judgment into a mathematical optimization problem: choose the interaction topology that minimizes the computation×regret product, and let the competitive allocation rule discover the cost structure that the topology makes observable. The designer's role is not to specify the optimal allocation—which is impossible without knowing future tasks—but to specify the *institution* within which the optimal allocation will be discovered through competition. The regret bounds tell us how quickly this discovery occurs, and the topology analysis tells us which institutions are most efficient.